\documentclass[sigconf]{acmart}

\AtBeginDocument{%
  \providecommand\BibTeX{{%
    \normalfont B\kern-0.5em{\scshape i\kern-0.25em b}\kern-0.8em\TeX}}}

\setcopyright{rightsretained}

\copyrightyear{2019}
\acmYear{2019}
\acmConference[MM '19]{Proceedings of the 27th ACM International Conference on Multimedia}{October 21--25, 2019}{Nice, France}
\acmBooktitle{Proceedings of the 27th ACM International Conference on Multimedia (MM '19), October 21--25, 2019, Nice, France}\acmDOI{10.1145/3343031.3350591}
\acmISBN{978-1-4503-6889-6/19/10}

\begin{document}
\fancyhead{}
\title{BUDA.ART: A Multimodal Content-Based Analysis \\and Retrieval System for Buddha Statues}

\author{Benjamin Renoust, Matheus Oliveira Franca, Jacob Chan, Van Le, Ayaka Uesaka, 
\mbox{Yuta Nakashima}, Hajime Nagahara}
\email{renoust@ids.osaka-u.ac.jp}
\affiliation{%
  \institution{Institute for Datability Science, Osaka University}
  \city{Osaka}
  \country{Japan}
}

\author{Jueren Wang}
\author{Yutaka Fujioka}
\email{fujioka@let.osaka-u.ac.jp}
\affiliation{%
  \institution{Graduate School of Letters, Osaka University}
  \city{Osaka}
  \country{Japan}}

\renewcommand{\shortauthors}{Renoust, et al.}

\begin{abstract}
 We introduce BUDA.ART, a system designed to assist researchers in Art History, to explore and analyze an archive of pictures of Buddha statues. The system combines different CBIR and classical retrieval techniques to assemble 2D pictures, 3D statue scans and meta-data, that is focused on the Buddha facial characteristics. We build the system from an archive of 50,000 Buddhism pictures, identify unique Buddha statues, extract contextual information, and provide specific facial embedding to first index the archive. The system allows for mobile, on-site search, and to explore similarities of statues in the archive. In addition, we provide search visualization and 3D analysis of the statues.
\end{abstract}



\begin{teaserfigure}
  \includegraphics[width=\textwidth]{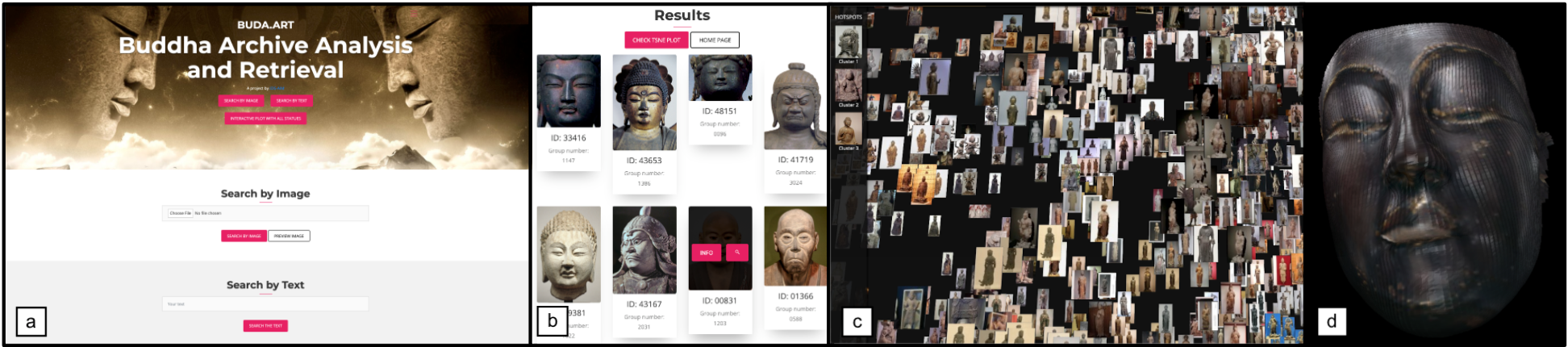}
  \caption{An overview of the BUDA.ART system, from search, to exploration and 3D visualization.}
  \label{fig:teaser}
\end{teaserfigure}

\maketitle

\section{Introduction}
The spread and evolution of Buddhism across Asia is the topic of many books~\cite{secret2006,faces2013}. Multiple theories are confronting on which path(s) this spread took place across the Asian subcontinent, reaching the coasts of the Japanese archipelago along the Silk Road~\cite{spread1986, Guide, secret2006,faces2013}. 
Buddhism brought many works of art and their rules so that local people would craft new artworks by themselves.
giving their identity to the resulting style~\cite{style1987}. 
Nowadays, only a few experts can identify these works \emph{subjective} to their own knowledge, sometimes disputing explanations~\cite{controversy2011}.

In order to investigate Buddhism at a large scale, we analyze a large archive of Buddhism related documents through the produced art. To do so, we focus on the representation of Buddha, which is central to Buddhism art. Although their exist statues of many different types, their construction respects \emph{canons}\footnote{A \emph{canon of art} refers to a universal set of rules and principles establishing the fundamentals and/or optimal.} which have been normalized over the centuries. Despite the rules, time and travels allowed for quite an evolution among the style of statues, and aligning many of these statues may allow us to capture the traces of this evolution~\cite{reportya}.

Using modern face detection and recognition~\cite{wang2018facial}, we focus on the faces of Buddha. Our experts have accumulated a large amount of photos and pictures related to Buddhism, that they cannot bring with them every time they visit a temple or museum, even less having an overview of their collection. 
Statues are 3D objects but 2D pictures usually poorly convey their spatial structure. So we build BUDA.ART (for \textbf{Bud}dha \textbf{A}rchive \textbf{A}naysis and \textbf{R}e\textbf{t}rieval, Fig.~\ref{fig:teaser}) a web-based system that combines and deliver all three aspects: knowledge/metadata, 2D pictures, and 3D structure, such that experts can query, search, and explore Buddha statues even on field. Such a retrieval system allows to explore a query space similarly to Barthel's map~\cite{barthel2017visually}, while it shares components of a CBIR~\cite{gomez2018demonstration}, it also provides hyperlinking as in for news video~\cite{eskevich2015hyper}, with new dedicated classifiers and 3D structure analysis for Buddha statues.

\section{Data preprocessing}

Our co-authors, experts in Art History from the Graduate School of Letters, have accumulated over 50,000 pictures of all kinds (about 500GB in total) in a semi-organized manner. They have been captured under many conditions, mainly: museum collection pieces acquired with standard methods, on-site captures in museum or temple, carefully captured Buddhist art treasures, outside field trips, and scans of dedicated literature. 

This is real data \emph{``in-the-wild''}: multiple size and formats; picture redundancy; not all pictures contain Buddha statues; a same statue can be taken across many angles and lights; multiple statues or subpictures in one picture; pictures can be a detail of larger artifact; this detail itself can be representation of Buddha. 
The pictures are not annotated (nor localized in EXIF), but they may be attached to indirect contextual information: it may be in their filename, or in their folders (over 1.7k different folders). We could extract some amount of contextual knowledge from this structure.

We first cleaned the dataset by removing near-duplicates using VGG16~\cite{simonyan2014very} embeddings, and used a t-SNE~\cite{maaten2008visualizing} 2D projection to visualize the space of the remaining pictures (~20k pictures, of which ~10k with a duplicate), so we could easily remove all non-relevant pictures. We created chains of pictures based on their creation order and cosine similarity to assign them to single Buddha statues. Similar (or duplicate) pictures were matched across chains (and folders) in an nearest neighbor graph to assemble same statues spread in different folders. A final round of manual annotation of this graph visualization allowed us to identify 3685 unique Buddha statues, among them we consider 804 statues with at least 5 pictures covering ~17k pictures. 

We extracted text content from 1.7k folder names and filenames, filtered locations and era when available, thus bringing context metadata when possible to the 804 statues.
We automatically recovered 366 statue with types, 672 with country, 461 with regions, and 460 with cities in which the statues were taken, 113 with countries, 104 with regions and 102 with cities of origin of the statues, 98 with construction eras, and 89 with temples.

For this archive, we first mined faces with the Faster R-CNN~\cite{sun2018face}, and manually corrected all annotations to form a ground truth of 1847 face pictures for the selected statues, that we used to fine tune the faster R-NN model (included RPN + VGG16) initially trained on Imagenet~\cite{deng2009imagenet}. We additionally use the VGGFace2~\cite{cao2018vggface2} trained ResNet50~\cite{he2016deep} model fine-tuned with our ground truth to compute face embedding.








\begin{figure}
    \centering
    \includegraphics[width=\linewidth]{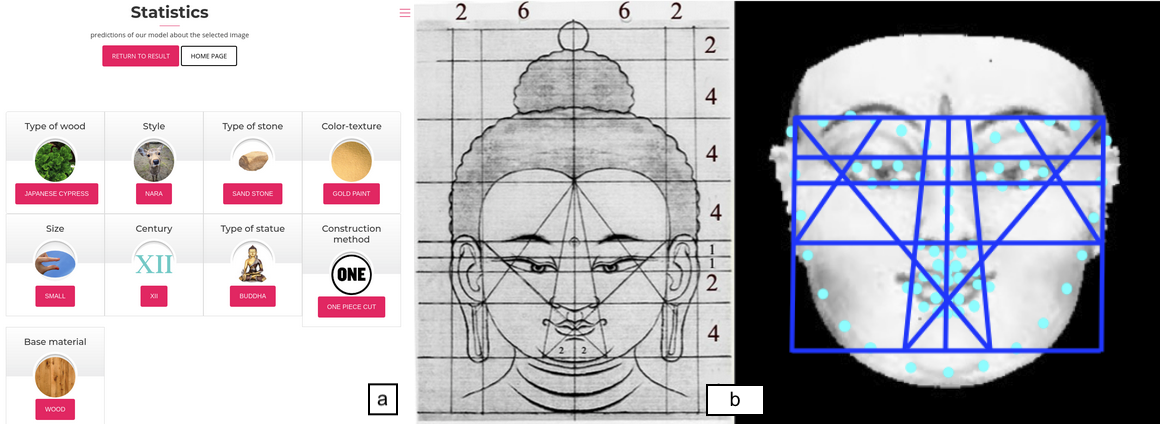}
    \caption{Label prediction and structure analysis.\vspace{-0.3cm}}
    \label{fig:stucture}
\end{figure}

\section{Retrieval and analysis}

We may now build the retrieval system to support experts when confronting statues. It is based on top of elasticsearch~\cite{divya2013elasticsearch} for text-based search and exploration (that is full text search from our extracted metadata, image path, and other user defined properties), and an image similarity search, as proposed by FAISS~\cite{JDH17} for image search and comparison. To further extend search, two types of embeddings are used: full image embeddings with Imagenet-trained VGG16~\cite{simonyan2014very} help to search pictures from their global information (for example, statues of similar shapes); face-dedicated embeddings with our fine-tuned ResNet50~\cite{he2016deep} to search for statues by their face and propose label prediction from our classifier~\cite{renoust2019historical} (Fig.~\ref{fig:stucture}a).

Users may then search statues by text or image and face (Fig.~\ref{fig:teaser}, a,b). Any face present in the uploaded picture is automatically sear\-ched using our Faster R-CNN model~\cite{sun2018face}. Users can additionally manually input a search area. Individual access to the results pre\-sents all the properties of a statues, all other pictures, such that experts may even edit this information. Each element of results  (image or text) is hyperlinked so it may become a new search element and support user exploratory search.

A neighborhood map is also created for users to explore the full content of the database (Fig.~\ref{fig:teaser}, c), as well as the neighborhood of a search result. This neighborhood map is built on top of a UMAP~\cite{mcinnes2018umap} projection of all image embeddings that constitute the search results (or all the database).

Acquiring 3D scans is not scalable, but our experts wish to investigate the 3D structure of the faces. We offer to interpolate Buddha faces in a 3D model using joint reconstruction and dense alignment~\cite{feng2018joint} (Fig.~\ref{fig:teaser}, d). The 3D model allows us to further explore facial landmarks, and historical facial proportions~\cite{renoust2019historical} (Fig.~\ref{fig:stucture}b).

The search system is built on a client-server architecture, on the server side, a python Flask~\cite{flask} framework encapsulates access to all the deep neural network models, elasticsearch~\cite{divya2013elasticsearch}, and FAISS~\cite{JDH17}. 
The neighborhood map is made using UMAP~\cite{mcinnes2018umap} and WebGL~\cite{webgl} with PixPlot~\cite{pixplot}. The 3D representation is built on top of three.js~\cite{threejs}. The web interface is built on top of UIdeck\cite{uideck}.
Our system is responsive so users can query it \textit{on-site} from a simple snapshot to a Buddha statue taken with any mobile device.








\section{Conclusion}

We presented a database construction down to search interface creation of 2D and 3D representations of Buddha statues. The system demonstrates easy on-line search of specific Buddha statues, following user defined criteria and different picture representation, with a query space visualization. We also demonstrate label prediction, 3D reconstruction of statues, comparison, and highlight structural feature. Experts can then query the database on-field with a simple smart phone. The upcoming future work will add a recommendation system based on nearest-neighbor search for each result. We will also deploy analysis of Buddha-statue specific 3D features on-the-fly, and add online comparison of 3D models.\\
\textbf{Acknowledgement:} Supported by JSPS KAKENHI \#18H03571. 


\bibliographystyle{ACM-Reference-Format}
\bibliography{bibliography}

\end{document}